# Q2A: Querying Implicit Fully Continuous Feature Pyramid to Align Features for Medical Image Segmentation

Jiahao Yu, and Li Chen

*Abstract*— Recent medical image segmentation methods apply implicit neural representation (INR) to the decoder for achieving a continuous coordinate decoding to tackle the drawback of conventional discrete grid-based data representations. However, the INR-based decoder cannot well handle the feature misalignment problem brought about by the naive latent code acquisition strategy in INR. Although there exist many feature alignment works, they all adopt a progressive multi-step aligning paradigm on a discrete feature pyramid, which is incompatible with the continuous one-step characteristics of INR-based decoder, and thus fails to be the solution. Therefore, we propose Q2A, a novel one-step query-based aligning paradigm, to solve the feature misalignment problem in the INR-based decoder. Specifically, for each target coordinate, Q2A first generates several queries depicting the spatial offsets and the cell resolutions of the contextual features aligned to the coordinate, then calculates the corresponding aligned features by feeding the queries into a novel implicit fully continuous feature pyramid (FCFP), finally fuses the aligned features to predict the class distribution. In FCFP, we further propose a novel universal partition-and-aggregate strategy (P&A) to replace the naive interpolation strategy for latent code acquisition in INR, which mitigates the information loss problem that occurs when the query cell resolution is relatively large and achieves an effective feature decoding at arbitrary continuous resolution. We conduct extensive experiments on two medical datasets, i.e. Glas and Synapse, and a universal dataset, i.e. Cityscapes, and they show the superiority of the proposed Q2A.

## I. Introduction

MEDICAL image segmentation is a fundamental technology for modern computer-assisted diagnosis (CAD) applications and image-guided surgery systems, which aims to segment objects of interest via pixel-wise classification on medical images. Modern medical image segmentation methods mainly adopt various neural network architectures to learn *discrete* grid-based data representations (e.g. rasterized label masks) [1]–[5]. Though effective, these methods have limited

Manuscript received 2 January 2024. This research was supported by the National Natural Science Foundation of China (Grant No.61972221). (*Corresponding author: Li Chen*.)
Jiahao Yu and Li Chen are with the School of Software, BNRist, Tsinghua University, Beijing 100084, China (e-mail: yujh21@mails.tsinghua.edu.cn; chenlee@tsinghua.edu.cn).

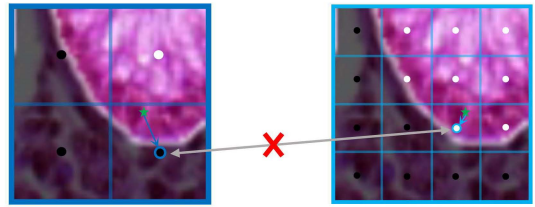

Fig. 1: An example of feature misalignment problem in implicit neural representation (INR). For the identical target coordinate (star), naive nearest neighbor interpolation obtains a latent code of background context (black circle) on $2 \times 2$ maps and a code of foreground context (white circle) on $4 \times 4$ maps respectively, where the context mismatch occurs.

spatial flexibility and poor computational scaling, and the pixel-wise supervision is unable to adequately model object boundaries [6], [7]. Therefore, some works [8]–[11] have explored to introduce *continuous* implicit neural representation (INR) to the decoder part of segmentation network, which formulates the segmentation map reconstruction process as a direct *one-step* coordinate signal decoding. This formulation, i.e. INR-based decoder, enables direct modeling of object contours, superior memory efficiency [12], and smooth predictions at arbitrary resolutions that are independent of input size.

However, all the existing INR-based decoders fail to well handle the feature misalignment problem [13], since at coordinate decoding phase these approaches obtain the coordinate-dependent latent code for INR by performing naive point interpolation, e.g. nearest neighbor, on multi-scale feature maps and the naive interpolation may lead to the context mismatch among the latent codes obtained from different scales, i.e. feature misalignment problem (Fig. 1). Although there exist many methods to cope with feature misalignment [13]–[17], they all fail to address the case of INR-based decoder, since these methods all employ a *multi-step* layer-wise aligning paradigm, which aligns features progressively on discrete scales pre-defined at the feature-pyramid-based decoder (Fig. 2a), and the multi-step style is not compatible with the one-step characteristics of INR-based decoder.

Accordingly, for the first time, this paper proposes a novel *one-step* query-based feature aligning paradigm (dubbed as Q2A) for INR-based decoder (Fig. 2b). Q2A consists of three primary modules, i.e. query generator, fully continuous feature pyramid (FCFP), and segmentation head. In the decoding process, for each target coordinate $\mathbf{p}$, the query generator first generates $K$ quadruple queries, depicting the spatial offsets and cell resolutions of the contextual features aligned to the target coordinate, then FCFP decodes the $K$ queries to directly obtain the $K$ aligned features in a one-step manner, and finally the aligned features are fused at the segmentation head to output the predicted distribution of the target coordinate. Specifically, the query generator and FCFP are modeled by INR, and the segmentation head is a typical position-wise multi-layer perceptron (MLP, a.k.a. $1 \times 1$ convolution). However, the design of FCFP requires some differences from the existing INR approaches [8]–[10], [18], because the query for FCFP contains cell resolution information and the default latent code acquisition strategy of INR, i.e.naive point interpolation, will cause severe information loss problem when the cell resolution of query is coarser than that of the given feature map (Fig. 3b).

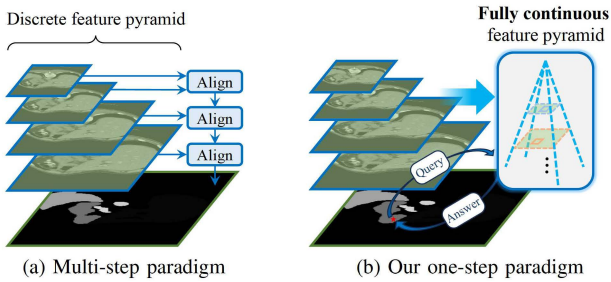

(a) Multi-step paradigm  (b) Our one-step paradigm

Fig. 2: Comparison of the proposed one-step feature aligning paradigm with the conventional multi-step one. a) Introduces layer-wise aligning modules to a typical discrete feature pyramid to perform feature aligning step-by-step. b) Our method directly queries a fully continuous feature pyramid built by an implicit neural field and answers the aligned contextual features to achieve the one-step feature aligning.

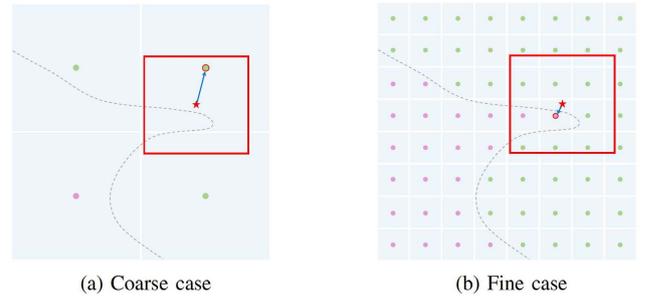

(a) Coarse case  (b) Fine case

Fig. 3: Two cases of the existing INR approaches utilizing naive nearest neighbor strategy for latent code acquisition to decode query, where the star and box indicate the spatial coordinate and cell resolution of a query respectively, pink circles and green ones represent latent codes with two different contexts, and the dotted gray curve indicates the continuous real-world boundary of two objects. a) When the rasterized map is coarser than the box, the obtained latent code can usually represent well the cell. b) When the rasterized map is finer, the default strategy obtains the code by referring to the nearest or the top four nearest codes (by bilinear interpolation) and ignores many other codes in the box, thus the obtained code is too local to represent the cell well, which is the information loss problem.

Thus, we further propose a novel universal partition-and-aggregate strategy (P&A) to address the information loss. Specifically, given a query cell and a feature map $F$ from the encoder, instead of performing point interpolation only at the center of the query cell on $F$, P&A first performs point interpolation at each subcells partitioned from the query cell on $F$, and then the subcell features are fused by voting-based aggregation to acquire the final latent code. P&A is essentially a universal latent code acquisition strategy, as it not only alleviates information loss when the resolution of the query is coarser than that of the given feature map but is also almost equivalent to naive point interpolation when the resolution of the query is finer. By integrating the P&A strategy into INR, for the first time, we model a *fully* continuous feature pyramid (FCFP) that can achieve effective feature decoding at arbitrary continuous resolution.

We demonstrate the effectiveness of Q2A with different segmentation architectures on two medical datasets, Glas [19] and Synapse [20]. We compare Q2A with some state-of-the-art feature alignment approaches and some vanilla INR-based decoders, and Q2A outperforms them in all cases. Furthermore, we conduct comparison experiments on Cityscapes [21] to validate the effectiveness of our Q2A on the universal segmentation scenario.

The major contributions can be summarized as follows.

- For the first time, we propose a novel *one-step* query-based feature aligning paradigm (dubbed as Q2A) based on implicit neural representation for medical image segmentation networks, which tackles the feature misalignment problem in the decoding process. This paradigm can also be applied to other dense prediction tasks with minor revision.
- We propose a novel universal partition-and-aggregate strategy (P&A) for latent code acquisition of implicit neural representation to model a *fully* continuous feature pyramid (FCFP) for the first time, where effective feature decoding can be achieved at arbitrary continuous resolution.
- We conduct extensive experiments on two medical datasets, Glas, Synapse, and a universal dataset, Cityscapes, and they demonstrate the superiority of the proposed Q2A.

## II. Related Work

### A. Medical Image Segmentation

Since the medical image segmentation task requires modeling the semantic context, scale information, and spatial relationships of the target object (e.g. organs, tumor), most methods adopt a multiscale scheme with an encoder-decoder architecture [22]–[25], where the encoder captures shallow spatial details and deep semantic information respectively in different layers and the decoder learns to fuse the feature maps of different scale levels. One seminal work is U-Net [1], based on which many approaches have been proposed [2], [26]–[28]. Considering the volumetric characteristics of medical images, some 2.5D approaches utilize tri-planar architectures to combine three-view slices for modeling each voxel [29]–[31], and some 2D&3D methods design different multi-dimensional feature fusion mechanisms to enrich the learned representations [27], [32]. Recently, some works introduced Transformer to reshape the architecture of encoders and decoders to better capture long-range dependencies [3]–[5], [33]–[35]. Despite the progress, the existing methods still suffer from two problems: 1) Most of them seek to learn *discrete* grid-based output representations, which results in spatial inflexibility and poor computational scaling. 2) Few works in the medical area specifically study the feature misalignment problem that occurs when multi-resolution feature maps are fused in the decoder. To deal with the two problems, our proposed Q2A exploits implicit neural field function to learn *continuous* representation and is tailored to tackle the feature misalignment problem.

### B. Implicit Neural Representation

Implicit neural representation (INR) aims to represent an object with a neural network (typically a multi-layer perceptron, i.e. MLP), which maps coordinates to signal. This paradigm was first employed to model object shape [36], [37], scene surface [38], [39], and structure appearance [40]–[42] in 3D reconstruction area, since INR maintains a small number of parameters in contrast to the explicit 3D representation, e.g. point cloud, mesh and voxel, and is also capable to capture the very fine details of the shape. Due to its efficiency and effectiveness, INR was further applied to 2D image super-resolution area [43]–[45], where the researchers mainly focused on reconstructing the high-frequency details of the images [46]–[48]. Recently, INR was introduced to the image segmentation decoder part to tackle the *discrete* characteristics of modern data representation [8]–[11], [18], [49], [50], whereas the straightforward application may bring the feature misalignment problem since the coordinate-dependent latent codes for INR are obtained by performing naive point interpolation on multi-resolution feature maps and this will lead to the context mismatch among the obtained codes. To address the problem, our Q2A builds a novel query-based feature aligning paradigm and designs a novel universal P&A strategy for code acquisition in INR.

### C. Feature Alignment

Feature alignment works study aligning multi-resolution feature maps at the decoder side of the segmentation network, as the spatial misalignment problem may occur when aggregating the late upsampled coarse features maps with the early high-resolution ones. A typical approach, i.e. U-Net [1], adds layer-wise skip connection paths to build a coarse-to-fine feature pyramid for encouraging the learning of context localization. However, the upsampling and the convolution operation cannot handle the feature misalignment problem well, since the upsampling only exploits relative spatial information and the convolution module is parameter-sharing. Although no prior works in the medical area specifically study the problem as far as we know, there are some aligning schemes in the semantic segmentation area. Specifically, Gated-SCNN [17] adopted a gated mechanism to adaptively fuse low-level details with high-level context. SFNet [14] and AlignSeg [13] introduced layer-wise aligning flow or transformation offsets to align the multi-level features. CARAFE [15], [16] designed a plug-in upsampling module with context-aware parameters to adaptively assemble features. Despite their effectiveness, they all essentially design a *multi-step* layer-wise aligning mechanism upon the feature-pyramid-based decoder, which is not compatible with the *one-step* characteristics of INR-based decoder. Differently, our Q2A is tailored for an INR-based decoder to accomplish a *one-step* feature aligning.

### D. Q2A vs. DeformableNet.

DeformableNets design deformable convolution [51], [52] or deformable attention [53], [54] to solve the misalignment problems in the vision tasks, which are superficially similar to our work. We emphasize that our Q2A and their works have at least three key differences. 1) *Different motivations:* We aim to adaptively align feature maps of different resolutions during aggregation, but they aim to learn the receptive field sizes for different objects. 2) *Different architectures:* Our Q2A displays a direct one-step architecture for decoding, but their deformable modules are layer-wise plug-ins of multi-step architectures. 3) *Different scale ranges:* Our query map is defined at an arbitrary continuous scale, but their offset maps are defined at specific discrete scales.

## III. METHOD

### A. Preliminary

This section presents our method via 2D case. Given a medical image $I \in \mathbb{R}^{C_I \times H \times W}$, medical image segmentation task aims to predict the corresponding segmentation map $P \in \mathbb{R}^{N \times H \times W}$, where $H$, $W$ and $C_I$ are the height, width and channel of the input image $I$, and $N$ denotes the number of target classes. To fulfill this task, an encoder-decoder paradigm is typically adopted, where the encoder gradually extracts the multiscale feature maps as the input image as forward propagated through the backbone network and the decoder further fuses the multiscale feature maps in a coarse-to-fine manner to produce the target segmentation maps. This paper focuses on the design of the decoder side. Following the setting in [55], we denote the multiscale feature maps from the encoder as $\{F_i\}_{i=2}^{5}$, where $F_i \in \mathbb{R}^{C_i \times \frac{H}{2^i} \times \frac{W}{2^i}}$ is the feature map of $i$-th level, $C_i$ is the corresponding channel dimension, and a larger $i$ denotes a deeper location in the encoder. Typical decoder architectures [1], [55] adopt a feature pyramid to fuse coarse-scale feature maps with fine-scale features in a top-down manner step by step. To better handle the feature misalignment problem, the existing solutions [13]–[16] all introduce layer-wise aligning modules to plug in the typical feature pyramid. Though effective, these methods cannot handle the case of the recent emerging INR-based decoder, since the step-by-step

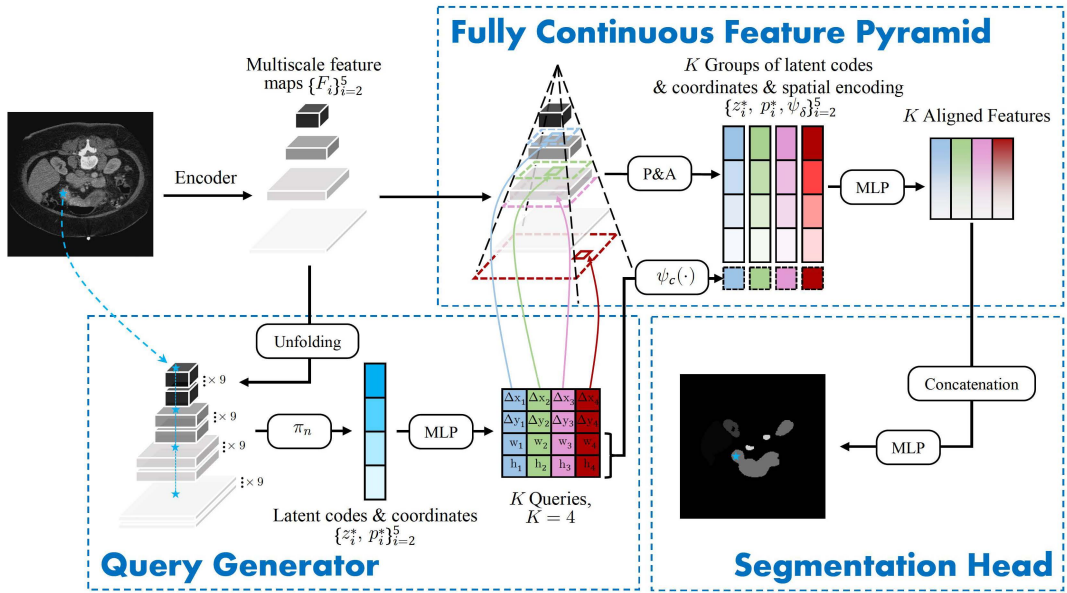

Fig. 4: The pipeline of the proposed Q2A. Given the multiscale feature maps extracted from the input image by an encoder, our Q2A proposes three modules, i.e. query generator $\mathcal{G}$, fully continuous feature pyramid $\mathcal{F}$, segmentation head $\mathcal{H}$, to output class distribution for arbitrary continuous coordinate. For a target coordinate(star), first $\mathcal{G}$ performs naive point interpolation $\pi_n$ on the unfolded multiscale feature maps to obtain latent codes and their coordinates, and feeds them to an MLP to acquire $K$ quadruple queries. Then the queries are fed to $\mathcal{F}$, and the novel P&A is performed to obtain $K$ groups of latent codes, their coordinates, and corresponding spatial encoding, which are further fed to an MLP along with the proposed cell embedding $\psi_c$ to obtain $K$ aligned features. Finally, $\mathcal{H}$ concatenates the aligned features and uses an MLP to predict class distribution.

aligning is not compatible with the continuous characteristics of INR.

### B. Overview of Q2A

In typical implicit neural representation (INR) ( [10], [44]), each image $I$ is represented as multiscale feature maps, i.e. $\{F_i\}_{i=2}^5$ extracted from the encoder, and a shared decoding function is defined to produce the continuous signal map $P$ over all the 2D discrete multiscale feature maps, where feature vectors are viewed as latent codes evenly distributed in the 2D space. Given a 2D continuous coordinate $\mathbf{p} \in \mathbb{R}^2$, the signal value is typically defined as

$$\text{INR}(\mathbf{p}, \mathbf{F}; \Theta) = f_\Theta \Big( \{z_i^*(\mathbf{p}; \pi_n), \mathbf{p} - p_i^*(\mathbf{p}; \pi_n)\}_{i=2}^5 \Big) \quad (1)$$

where $\mathbf{F}$ denotes multiscale features $\{F_i\}_{i=2}^5$, $f_\Theta$ is an MLP and $\Theta$ are its parameters, $z_i^*(\mathbf{p}; \pi_n)$ denotes the latent code obtained by strategy $\pi_n$ from coordinate $\mathbf{p}$ on the $i$-th feature map, $p_i^*(\mathbf{p}; \pi_n)$ is the coordinate of the obtained latent code, and strategy $\pi_n$ denotes nearest neighbor interpolation. Note that all continuous 2D coordinates in this section are defined in $[-1, 1] \times [-1, 1]$ and can be scaled up to any discrete resolution to obtain the latent code on the given feature maps. Unfortunately, this typical scheme suffers from the feature misalignment problem, since the naive strategy $\pi_n$ may result in context mismatch among the latent codes obtained from different scales.

To tackle this problem, we propose Q2A, which feeds an implicit fully continuous feature pyramid with **Q**ueries to **A**lign features for each target coordinate. Different from the conventional INR, Q2A consists of three decoding modules, i.e. query generator $\mathcal{G}$, fully continuous feature pyramid $\mathcal{F}$ (FCFP), and segmentation head $\mathcal{H}$. Fig. 4 shows the pipeline of our Q2A for medical image segmentation. In the decoding process, for each target coordinate $\mathbf{p} \in \mathbb{R}^2$, the query generator $\mathcal{G}$ first produces $K$ quadruple queries $\{Q(\mathbf{p}, k)\}_{k=1}^K$, where query $Q(\mathbf{p}, k) \in \mathbb{R}^4$ depicts the spatial offsets and cell resolutions of the $k$-th contextual features aligned to $\mathbf{p}$. Then FCFP $\mathcal{F}$ decodes the $K$ queries to directly obtain the $K$ aligned features $\{A(\mathbf{p}, k)\}_{k=1}^K$ in a one-step manner. Finally, the aligned features are fused at the segmentation head $\mathcal{H}$ to output the predicted distribution $\hat{y} \in \mathbb{R}^N$ of the target coordinate. Specifically, query generator $\mathcal{G}$ and FCFP $\mathcal{F}$ are modeled by INR respectively, and head $\mathcal{H}$ is simply an MLP. At the inference stage, to predict a discrete segmentation map $P$ from the multiscale representation $\{F_i\}_{i=2}^5$, first a query map is produced by $Q = \mathcal{G}(\{F_i\}_{i=2}^5)$, $Q \in \mathbb{R}^{4K \times H_q \times W_q}$ where $H_q$, $W_q$ are the height and width of the query map $Q$ and they also determine the spatial size of the segmentation result $P$. Then the aligned features are obtained by $A = \mathcal{F}(\{F_i\}_{i=2}^5, Q)$, $A \in \mathbb{R}^{C_a K \times H_q \times W_q}$ where $C_a$ is the channel of each aligned feature. Finally, the segmentation results are generated by $P = \mathcal{H}(A)$, $P \in \mathbb{R}^{N \times H_q \times W_q}$.

In the following parts, we first present the query generator in Sec. III-C, then we display FCFP and the proposed P&A in Sec. III-D.

### C. Query Generator

For a continuous coordinate $\mathbf{p} \in \mathbb{R}^2$, our query generator $\mathcal{G}$ will produce $K$ quadruple queries, and each query can be decomposed as

$$Q(\mathbf{p}, k) = \{\Delta\mathrm{x}_k(\mathbf{p}), \Delta\mathrm{y}_k(\mathbf{p}), \mathrm{w}_k(\mathbf{p}), \mathrm{h}_k(\mathbf{p})\} \quad (2)$$

where $Q(\mathbf{p}, k) \in \mathbb{R}^4$ denotes the $k$-th query for coordinate $\mathbf{p}$, $\Delta\mathrm{x}_k(\mathbf{p})$ denotes coordinate offset on width axis between the $k$-th aligned feature $A(\mathbf{p}, k)$ and target coordinate $\mathbf{p}$, $\Delta\mathrm{y}_k(\mathbf{p})$ denotes offset on height axis between the $k$-th aligned feature $A(\mathbf{p}, k)$ and target coordinate $\mathbf{p}$, $\mathrm{w}_k(\mathbf{p})$ denotes the width of pixel cell of the $k$-th aligned feature $A(\mathbf{p}, k)$, and $\mathrm{h}_k(\mathbf{p})$ denotes the height of pixel cell of the $k$-th aligned feature $A(\mathbf{p}, k)$. We further define $\delta_k(\mathbf{p}) = \{\Delta\mathrm{x}_k(\mathbf{p}), \Delta\mathrm{y}_k(\mathbf{p})\}$ and $c_k(\mathbf{p}) = \{\mathrm{w}_k(\mathbf{p}), \mathrm{h}_k(\mathbf{p})\}$, where $\delta_k(\mathbf{p}) \in \mathbb{R}^2$ represents spatial offset between the $k$-th aligned feature $A(\mathbf{p}, k)$ and coordinate $\mathbf{p}$, and $c_k(\mathbf{p}) \in \mathbb{R}^2$ represents cell decoding resolution of the $k$-th aligned feature $A(\mathbf{p}, k)$. The following content introduces the design of our query generator.

*1) Feature unfolding:* Since the query generator is responsible for generating the offsets, which are related to the neighborhood information of the target coordinate, it probably effectively improves the quality of the generated query that expanding the receptive field of the coordinate-dependent latent code. Thus we follow [44] to apply feature unfolding to $\mathbf{F}$ and get $\hat{\mathbf{F}}$. Each latent code in $\hat{\mathbf{F}}$ is the concatenation of the $3 \times 3$ neighboring latent codes in $\mathbf{F}$. Mathematically, the feature unfolding is defined as

$$\hat{\mathbf{F}}^{(i)}_{j,k} = \oplus\big(\{\mathbf{F}^{(i)}_{j+l,k+m}\}_{l,m\in\{-1,0,1\}}\big) \quad (3)$$

where $\hat{\mathbf{F}}^{(i)}_{j,k}$ denotes the feature value of the $j$-th row and the $k$-th column of the $i$-th feature map in $\hat{\mathbf{F}}$, and $\oplus(\cdot)$ denotes the concatenation of a set of vectors.

*2) Generator design:* Based on (1), given a coordinate $\mathbf{p}$, a quadruple query is produced by

$$\Delta\mathrm{x}_k(\mathbf{p}) = \sigma\big(\mathrm{INR}(\mathbf{p}, \hat{\mathbf{F}}; \Theta_{\mathcal{G}}^{\Delta\mathrm{X}})\big) \quad (4)$$

$$\Delta\mathrm{y}_k(\mathbf{p}) = \sigma\big(\mathrm{INR}(\mathbf{p}, \hat{\mathbf{F}}; \Theta_{\mathcal{G}}^{\Delta\mathrm{Y}})\big) \quad (5)$$

$$\mathrm{w}_k(\mathbf{p}) = \exp\big(\tau_1^w + \tau_2^w \times \sigma\big(\mathrm{INR}(\mathbf{p}, \hat{\mathbf{F}}; \Theta_{\mathcal{G}}^{\mathrm{W}})\big)\big) \quad (6)$$

$$\mathrm{h}_k(\mathbf{p}) = \exp\big(\tau_1^h + \tau_2^h \times \sigma\big(\mathrm{INR}(\mathbf{p}, \hat{\mathbf{F}}; \Theta_{\mathcal{G}}^{\mathrm{H}})\big)\big) \quad (7)$$

where $\sigma$ is $\tanh$ function, $\tau_1^w, \tau_2^w, \tau_1^h$, and $\tau_2^h$ are hyperparameters. The utilization of exponential functions $\exp(\cdot)$ in (6),(7) aims to encourage the generator to regulate the cell resolution of queries at the exponential level. In implementation, these four MLPs, i.e. $f_{\Theta_{\mathcal{G}}^{\Delta\mathrm{X}}}$, $f_{\Theta_{\mathcal{G}}^{\Delta\mathrm{Y}}}$, $f_{\Theta_{\mathcal{G}}^{\mathrm{W}}}$, and $f_{\Theta_{\mathcal{G}}^{\mathrm{H}}}$, share all parameters except for the last fully connected layer. Under this design, the value ranges of the generated query are $\delta_k(\mathbf{p}) \in (-1, 1) \times (-1, 1)$ and $c_k(\mathbf{p}) \in (\exp(\tau_1^w - \tau_2^w), \exp(\tau_1^w + \tau_2^w)) \times (\exp(\tau_1^h - \tau_2^h), \exp(\tau_1^h + \tau_2^h))$.

### D. Fully Continuous Feature Pyramid

Since a conventional feature pyramid is defined with pre-defined discrete scales (resolutions), it is unable to support the decoding of queries from $\mathcal{G}$, which is continuous at scale. Thus we aim to define a continuous feature pyramid (CFP), which supports feature decoding at arbitrary continuous scale and coordinate. Inspired by [10], given a coordinate $\mathbf{p}$ and its $k$-th query $Q(\mathbf{p}, k)$, a feature decoded from CFP is defined as

$$\mathrm{CFP}(\mathbf{p}, Q(\mathbf{p}, k), \mathbf{F}; \Theta_{\mathcal{F}}) = f_{\Theta_{\mathcal{F}}}\Big(\big\{z_i^*(\hat{\mathbf{p}}_k; \pi_n), \\ \hat{\mathbf{p}}_k - p_i^*(\hat{\mathbf{p}}_k; \pi_n), \psi_\delta(\hat{\mathbf{p}}_k - p_i^*(\hat{\mathbf{p}}_k; \pi_n))\big\}_{i=2}^5\Big) \quad (8)$$

where $\mathbf{F}$ denotes multiscale feature maps, $f_{\Theta_{\mathcal{F}}}$ is a MLP and $\Theta_{\mathcal{F}}$ are its parameters, coordinate $\hat{\mathbf{p}}_k = \mathbf{p} + \delta_k(\mathbf{p})$, and $\psi_\delta(\cdot)$ denotes spatial encoding. Different from (1), CFP further incorporates spatial encoding to facilitate feature decoding ability, which is because feature decoding task is essentially data reconstruction and previous works [40], [48] show that naive representation based on $xy$ coordinates (e.g. $\hat{\mathbf{p}}_k - p_i^*(\hat{\mathbf{p}}_k; \pi_n)$) fails to capture high-frequency signals and results in inferior reconstruction ability of network. Accordingly we follow [48] to introduce spatial encoding to (8), which is defined as

$$\psi_\delta(x) = \big\{\sin(\omega_1 x), \cos(\omega_1 x), \ldots, \sin(\omega_L x), \cos(\omega_L x)\big\} \quad (9)$$

where $\omega_l$ are initialized as $\omega_l = 2^l, l \in \{1, 2, \ldots, L\}$ and can be finetuned during training.

However, the design of CFP (8) has two problems. First, naive $\pi_n$ will bring severe information loss problem(Fig. 3b) when the resolution $c_k(\mathbf{p})$ is coarser than that of the given feature maps (e.g. the finest $F_2$ in $\mathbf{F}$). Second, the output result of (8) is independent of cell resolution information $c_k(\mathbf{p})$. These two problems both impair the desirable *continuous* characteristics of CFP, which is that effective feature decoding can be achieved at arbitrary continuous scale and coordinate. To address them, we propose a novel partition-and-aggregate strategy (P&A) to replace $\pi_n$, and further introduce cell embedding, to build a *fully* continuous feature pyramid $\mathcal{F}$ (FCFP).

*1) Partition-and-aggregate strategy (P&A):* Given a query coordinate $\hat{\mathbf{p}}_k$ and its resolution $c_k(\mathbf{p})$, our P&A aims to calculate a latent code on each feature map from $\mathbf{F}$, which can better represent the information in the query cell area. To achieve this goal, P&A first equally partitions the query cell into $s \times s$ subcells and employs $\pi_n$ for each subcell center to obtain $s \times s$ subcell latent codes, then P&A aggregates them to get the final latent code, where a simple average aggregation is adopted. Mathematically, P&A for latent code is defined as

$$z_i^*(\hat{\mathbf{p}}_k, c_k(\mathbf{p}); \pi_{pa}) = \frac{1}{s^2} \sum_{1 \leq j, l \leq s} z_i^*(\hat{\mathbf{p}}_k + \delta_{j,l}(c_k(\mathbf{p})); \pi_n) \quad (10)$$

where $\pi_{pa}$ denotes P&A, $\delta_{j,l}(c_k(\mathbf{p}))$ denotes coordinate offsets between $\hat{\mathbf{p}}_k$ and the center of the $j$-th row, $l$-th column subcell. Though effective for latent features, simply averaging the corresponding latent code coordinates will bring a naive coordinate result, thus an alternative aggregation function is in demand, which aims to enrich the representation of the coordinate of $z_i^*(\hat{\mathbf{p}}_k, c_k(\mathbf{p}); \pi_{pa})$ and is restricted to satisfying the permutation-invariant characteristics. Thus, we propose a voting-based aggregation function, which projects the distance relationship among the $s \times s$ subcell latent codes in the high-dimensional feature space onto the 2D $xy$ plane. Specifically, under a query cell with $s \times s$ subcells, the vote from the $x$-th subcell to the $y$-th subcell is defined as $\varrho(x,y) = \exp(-\|z_x^* - z_y^*\|_2)$, $1 \leq x, y \leq s^2$, where $z_x^*$ denotes the latent code of the $x$-th subcell center obtained by $\pi_n$ and $\|\cdot\|$ is Euclidean norm. $\varrho(x,y)$ satisfies commutative symmetry and implies that a feature pair with more similarity brings larger votes for each other. Then the total votes received by the $j$-th row, $l$-th column subcell is defined as $\alpha_{j,l} = \prod_{1 \leq u \leq s^2} \varrho(u, (j-1) \times s + l)$, and we normalize the total votes to obtain the weights for aggregation by $\hat{\alpha}_{j,l} = \alpha_{j,l}/\sum_{1 \leq u,v \leq s} \alpha_{u,v}$. Finally, P&A calculates the coordinate of the final latent code as

$$p_i^*(\hat{\mathbf{p}}_k, c_k(\mathbf{p}); \pi_{pa}) = \sum_{1 \leq j,l \leq s} \hat{\alpha}_{j,l} \times p_i^*(\hat{\mathbf{p}}_k + \delta_{j,l}(c_k(\mathbf{p})); \pi_n)$$
$$\sum_{1 \leq j,l \leq s} \hat{\alpha}_{j,l} = 1 \quad (11)$$

Equation (11) essentially pushes the 2D result coordinate closer to the subcell latent code which is the closest to the latent feature centroid, and thus it provides a better representation of the coordinate of $z_i^*(\hat{\mathbf{p}}_k, c_k(\mathbf{p}); \pi_{pa})$. In implementation, we apply `stop_grad` function to $p_i^*(\hat{\mathbf{p}}_k, c_k(\mathbf{p}); \pi_{pa})$ for stopping its gradient back-propagation in backward phase, which avoids extra computational costs in the training phase. Based on (10) and (11), our P&A is essentially a universal latent code acquisition strategy for INR (Fig. 5), which is because it not only alleviates the information loss problem when resolution $c_k(\mathbf{p})$ is coarser than that of the given map, but also degenerates into naive point interpolation when $c_k(\mathbf{p})$ is finer, whose effectiveness has been proven by previous works [9], [10], [18].

*2) Cell embedding:* Although P&A indirectly makes the decoding result dependent of the query resolution $c_k(\mathbf{p})$, our $\mathcal{F}$ still cannot directly perceive this information. Therefore we introduce cell embedding to enhance the network awareness of the cell resolution information, which is defined as

$$\psi_c(x) = x\mathbf{e} \quad (12)$$

where $\mathbf{e}$ is initialized as $\mathbf{e} = \mathbf{1}^{C_e} \in \mathbb{R}^{C_e}$ ($C_e$ is a hyperparameter) and can be finetuned during training.

*3) Implicit FCFP:* With the help of P&A and cell embedding, the implicit FCFP function $\mathcal{F}$ is finally defined as

$$\text{FCFP}(\mathbf{p}, Q(\mathbf{p}, k), \mathbf{F}; \Theta_\mathcal{F}) = f_{\Theta_\mathcal{F}}\Big(\big\{z_i^*(\hat{\mathbf{p}}_k, c_k(\mathbf{p}); \pi_{pa}),$$
$$\hat{\mathbf{p}}_k - p_i^*(\hat{\mathbf{p}}_k, c_k(\mathbf{p}); \pi_{pa}),$$
$$\psi_\delta(\hat{\mathbf{p}}_k - p_i^*(\hat{\mathbf{p}}_k, c_k(\mathbf{p}); \pi_{pa}))\big\}_{i=2}^5, \psi_c(c_k(\mathbf{p}))\Big) \quad (13)$$

As far as we know, we are the first to model a *fully* continuous feature pyramid (FCFP) where effective feature decoding can be achieved at arbitrary continuous resolution. Therefore, we

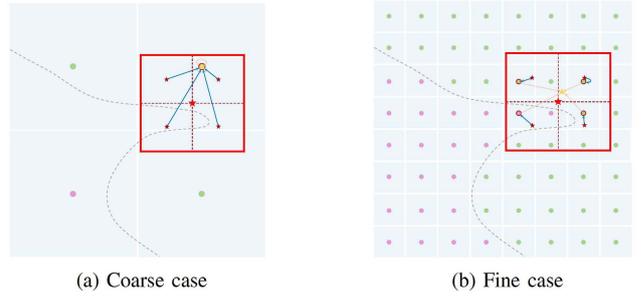

(a) Coarse case  (b) Fine case

Fig. 5: Same two cases as in Fig. 3, where we utilize P&A strategy for latent code acquisition. The small stars indicate subcells centers ($s = 2$), the arrow indicates nearest neighbor interpolation, the dotted arrow indicates voting-based aggregation, the triangle represents coordinate of the final latent code, and the other settings follow Fig. 3. a) When the rasterized map is coarser than the box, the obtained latent code is identical to the one obtained by nearest neighbor interpolation. b) When the rasterized map is finer, P&A provides a boarder view of the cell and the aggregated latent code can better represent the query cell.

obtain the $k$-th contextual feature aligned to target coordinate $\mathbf{p}$ by

$$A(\mathbf{p}, k) = \text{FCFP}(\mathbf{p}, Q(\mathbf{p}, k), \mathbf{F}; \Theta_\mathcal{F}) \quad (14)$$

## IV. EXPERIMENTS

### A. Datasets and Evaluation Metrics

*1) Datasets:* a) Glas [19] is a colon histology image dataset for gland segmentation, which provides 165 images of hematoxylin and eosin (H&E)-stained slides. Each pixel belongs to one of two classes. Each image is of $512 \times 512$ resolution. Following [56], the 165 images are split into 85 images for training and 80 for testing. b) Synapse [20] is a clinical CT image dataset for multi-organ segmentation, which contains 30 contrast-enhanced CT scans in 8 abdominal organs, with 3779 axial CT images in total. Each CT volume consists of 85~198 slices of $512 \times 512$ resolution, with a voxel spatial resolution of $0.54 \times 0.98 \times [2.5 \sim 5.0]\,\text{mm}^3$. Following [33], we use a random split of 18 training cases (2212 axial slices) and 12 cases for validation. c) Cityscapes [21] is a high-resolution urban scene dataset for semantic segmentation task, which contains 5000 finely annotated images and 20000 coarsely annotated images. Each image is of $1024 \times 2048$ resolution. Each pixel belongs to one of 19 classes. Following [10], [13], we only use the 5000 finely annotated images for experiments, which are further split into 2975, 500 and 1525 images for training, validation, and testing respectively.

*2) Evaluation metrics:* For Glas, we use average dice score and average Hausdorff distance as metrics. For Synapse, we adopt class-wise dice score averaged over cases and class-wise Hausdorff distance averaged over cases as metrics for multi-organ segmentation. For Cityscapes, we adopt Intersection over Union averaged over all classes (mIoU) as a metric. The number of float-point operations (FLOPs) and the number of parameters are also employed for efficiency evaluations.

| Method | Glas | | | | Synapse | | | | Cityscapes | | | |
|---|---|---|---|---|---|---|---|---|---|---|---|---|
| | #Params | GFLOPs | Dice(%) | HD95(mm) | #Params | GFLOPs | Dice(%) | HD95(mm) | #Params | GFLOPs | mIoU(%)(SS/MS) | |
| Bilinear Up-sampling | 11.24M | 10.42 | 87.98 | 18.31 | 15.31M | 16.11 | 73.24 | 39.06 | 27.7M | 183.4 | 74.47 | 76.46 |
| Nearest Neighbor | 11.24M | 10.42 | 87.02 | 18.65 | 15.31M | 16.11 | 72.44 | 39.23 | 27.7M | 183.4 | 75.83 | 77.03 |
| Deconvolution [57] | 12.93M | 16.33 | 87.50 | 18.15 | 16.29M | 26.48 | 73.21 | 37.08 | 29.5M | 304.4 | 71.03 | 72.70 |
| Semantic-FPN [55] | - | - | - | - | - | - | - | - | 31.0M | 219.1 | 78.00 | 79.14 |
| UNet [1] | 14.33M | 10.46 | 88.99 | 18.51 | 17.26M | 30.66 | 74.68 | 36.87 | - | - | - | - |
| CARAFE++ [16] | 13.81M | 24.25 | 91.64 | 11.75 | 17.46M | 41.88 | 76.79 | 29.02 | 30.3M | 211.5 | 77.50 | 78.62 |
| SFNet [14] | 22.40M | 51.51 | 92.17 | 9.33 | 19.77M | 177.73 | 77.92 | 29.81 | 42.9M | 327.3 | 79.14 | 79.67 |
| AlignSeg [13] | 19.13M | 68.21 | 92.31 | 7.99 | 21.61M | 194.78 | 77.42 | 29.60 | 49.7M | 348.6 | 78.72 | 79.92 |
| IFA [10] | 11.79M | 16.00 | 90.67 | 12.52 | 15.51M | 24.32 | 76.23 | 32.54 | 27.8M | 186.9 | 78.04 | 79.35 |
| IOSNet [9] | - | - | - | - | 15.49M | 23.08 | 75.56 | 29.17 | - | - | - | - |
| **Q2A(Ours)** | 11.49M | 18.44 | **93.29** | **5.48** | 17.49M | 35.19 | **78.99** | **24.93** | 27.9M | 258.7 | **79.27** | **80.16** |

TABLE I: Performance comparisons of different aligning methods on Glas test set, Synapse val set, and Cityscapes `val` set, respectively. GFLOPs calculations employ $512 \times 512$, $224 \times 224$, and $1024 \times 1024$ images for Glas, Synapse, and Cityscapes, respectively. 'SS' and 'MS' denote single-scale and multiscale inference respectively.

### B. Implementation Details

We implement all models with PyTorch toolbox [58], and use one single NVIDIA RTX 3090 GPU for Glas and Synapse, and eight for Cityscapes. We use cross entropy loss and dice loss as the loss functions for Glas and Synapse, and use cross entropy loss with online hard example mining (OHEM) for Cityscapes. For Glas, we use Adam optimizer with an initial learning rate of 0.001 and adopt `ReduceLROnPlateau` scheduler in PyTorch with patience as 20 and factor as 0.1. For Synapse and Cityscapes, we use stochastic gradient descent (SGD) optimizer with an initial learning rate of 0.01, weight decay of 0.0005 and momentum of 0.9, and adopt the 'poly' learning rate policy, where the initial learning rate is multiplied by $(1 - \frac{\text{iter}}{\text{max\_iter}})^{0.9}$. For Glas, we adopt batch size as 16 and training epochs as 200. For Synapse, we adopt resize size as $224 \times 224$, batch size as 24, and training epochs as 150. For Cityscapes, We adopt crop size as $769 \times 769$, batch size as 16, and training iterations as 18k. We set $s$ to 2, $\tau_1^w = \tau_1^h$, $\tau_2^w = \tau_2^h$, and the MLP in query generator is a fully connected layer. For Glas and Synapse, $C_e, C_a, L$ are set to 2, 64, 2, respectively, and the MLP in FCFP is of three layers with hidden dimensions as 512 and 256. For Cityscapes, $C_e, C_a, L$ are set to 6, 256, and 6, respectively, and the MLP in FCFP is of four layers with hidden dimensions as 512, 256, and 256. $K$ is 4, 3, and 4 for Glas, Synapse, and Cityscapes, respectively. $(\tau_1^h, \tau_2^h)$ is set to $(-\frac{9}{2}\ln 2, \frac{5}{2}\ln 2)$, $(\ln\frac{2}{51}, 2\ln 2)$, $(\ln\frac{8\sqrt{2}}{769}, \frac{5}{2}\ln 2)$ for Glas, Synapse, and Cityscapes, respectively.

### C. Results

*1) Aligning method comparison:* We compare our Q2A against three groups of aligning methods, i.e. naive interpolation aligning (bilinear up-sampling and nearest neighbor), tailored aligning methods based on feature pyramid including state-of-the-art methods like AlignSeg [13] and SFNet [14], recent emerging INR-based methods (IFA [10], IOSNet [9]). Specifically, we adopt ResNet-18, FCN, and ResNet-50 pretrained on ImageNet, as backbone encoder for Glas, Synapse, and Cityscapes, respectively. As shown in Tab. I, our Q2A achieves the best performance over other baseline methods. On one hand, though AlignSeg and SFNet also obtain relatively high performance, their overheads brought to the decoder side are much higher than ours and their step-by-step aligning paradigm is incompatible with the INR-based decoder. On the other hand, our Q2A brings a large amount of improvement to interpolation methods and INR-based decoders along with only a few overheads. Thus, our Q2A further achieves a more superior trade-off between computational cost and accuracy than all the previous methods. We further visualize some examples in Fig. 6 to show the superiority of our Q2A.

| Method | Glas | | Synapse | |
|---|---|---|---|---|
| | Dice(%) | HD95(mm) | Dice(%) | HD95(mm) |
| Q2A w/o P&A | 92.74 | 12.83 | 78.12 | 31.93 |
| Q2A w/o $\psi_c(\cdot)$ | 92.81 | 11.67 | 78.29 | 31.70 |
| Q2A w/o Voting | 93.01 | 8.99 | 78.64 | 28.25 |
| Q2A w/o $\psi_\delta(\cdot)$ | 93.09 | 8.69 | 78.65 | 26.37 |
| Q2A w/o Unfolding | 93.14 | 7.74 | 78.80 | 25.64 |
| Q2A (full) | 93.29 | 5.48 | 78.99 | 24.93 |
| Q2A w/o `stop_grad` | **93.85** | **4.09** | **79.17** | **23.88** |

TABLE II: Ablation study of our Q2A on Glas test set and Synapse val set. $\psi_c(\cdot)$ denotes cell embedding in (12). 'Voting' denotes voting-based aggregation in (11). $\psi_\delta(\cdot)$ denotes spatial encoding in (9). 'Unfolding' denotes the feature unfolding in (3). `stop_grad` refers to the gradient blocking of (11).

*2) Ablation studies:* We study the detailed contribution of each component in our Q2A by conducting ablation experiments in Tab. II. It shows that our P&A strategy is the most vital part of our model, and thus effective for alleviating the information loss problem in conventional INR-based decoders. Tab. II also shows that cell embedding $\psi_c(\cdot)$ is critical for model performance as it enhances the model awareness of cell resolution information. The results also show that voting-based aggregation and spatial encoding $\psi_\delta(\cdot)$ have similar importance for reconstructing features in the FCFP module. Moreover, we further remove the `stop_grad` function applied to (11), and it brings an extra boost to the performance of our Q2A, which implies that `stop_grad` function is probably not indispensable when resources for training Q2A is adequate and in inference phase the ablated version consumes equivalent amount of computation to the vanilla one.

*3) Comparison with state-of-the-arts:* We further compare our Q2A with other state-of-the-art methods on medical

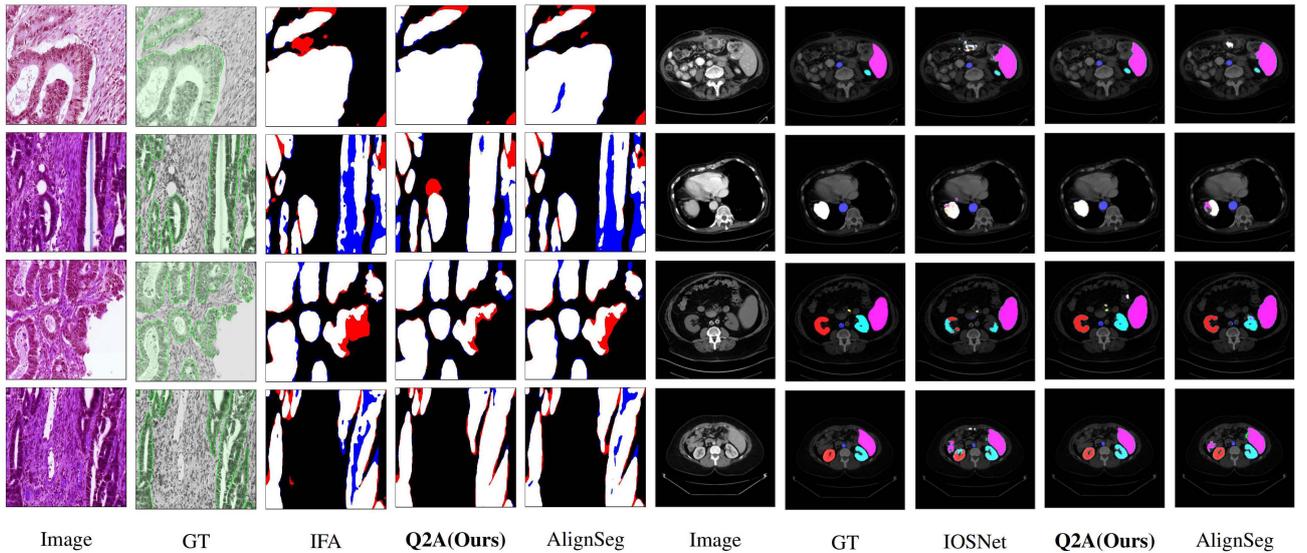

Fig. 6: Some visualization results on the Glas test set and Synapse val set. 'GT' indicates ground truth. In prediction results for Glas, white pixels represent positive, black pixels represent background (negative), red pixel represents false positive, and blue pixel represents false negative. The prediction results of the recent INR-based method (IFA [10], IOSNet [9]) suffer from severe contextual mismatch on a large number of locations, whereas our Q2A overcomes the mismatch problem by introducing feature aligning, which even achieves more superior performance than the state-of-the-art aligning method (AlignSeg [13]).

(a) Results on Glas.

| Method | Backbone | Dice(%) |
|---|---|---|
| UNet [1] | ResNet-18 | 88.99 |
| UNet++ [26] | ResNet-50 | 89.98 |
| CENet [22] | ResNet-34 | 89.02 |
| AttnUNet [24] | FCN | 87.68 |
| DeepLabv3 [59] | DResNet-50 | 87.49 |
| PraNet [25] | Res2Net-50 | 91.20 |
| SegFormer [60] | MiT-B2 | 89.73 |
| SETR-PUP [61] | T-Base | 88.75 |
| MedT [3] | GAT | 82.52 |
| Swin-UNet [35] | Swin-B | 88.94 |
| MCTrans [34] | ResNet-34 + MCT | 90.71 |
| TransUNet [33] | R50-ViT | 89.94 |
| UCTransNet [62] | ResNet-50 + CCT | 90.18 |
| TransFuse [4] | ResNet-34 + DeiT-S | 90.79 |
| ConTrans [5] | Swin-B + DAB-CNN | 92.06 |
| **Q2A(Ours)** | ResNet-18 | **93.29** |

(b) Results on Synapse.

| Method | Backbone | Dice(%) | HD95(mm) |
|---|---|---|---|
| UNet [1] | FCN | 74.68 | 36.87 |
| V-Net [28] | 3D FCN | 68.81 | - |
| UNet++ [26] | ResNet-50 | 76.91 | 36.93 |
| AttnUNet [24] | FCN | 75.57 | 36.97 |
| DARR [23] | 3D FCN | 69.77 | - |
| ResUNet [63] | ResUNet-a | 76.95 | 38.44 |
| MultiResUNet [64] | MultiRes-CNN | 77.42 | 36.84 |
| OSSNet [8] | 3D ResNet | 76.11 | - |
| TransUNet [33] | R50-ViT | 77.48 | 31.69 |
| TransNorm [65] | FCN + ViT | 78.40 | 30.25 |
| UCTransNet [62] | ResNet-50 + CCT | 78.23 | 26.75 |
| MT-UNet [66] | MTM | 78.59 | 26.59 |
| **Q2A(Ours)** | FCN | **78.99** | **24.93** |

(c) Results on Cityscapes.

| Method | Backbone | mIoU(%)(SS/MS) | |
|---|---|---|---|
| PSPNet [67] | ResNet-101 | 78.34 | 79.74 |
| DeepLabv3 [59] | DResNet-101 | 79.27 | 80.11 |
| DeepLabv3+ [68] | DResNet-101 | 79.46 | 80.50 |
| UPerNet [69] | ResNet-101 | 79.03 | 80.77 |
| CCNet [70] | ResNet-101 | 79.45 | 80.66 |
| SETR [61] | ViT-L | 79.21 | 81.02 |
| SFNet [14] | ResNet-101 | 79.80 | - |
| IFA [10] | ResNet-101 | 79.92 | 81.05 |
| **Q2A(Ours)** | ResNet-101 | **80.45** | **81.28** |

TABLE III: Comparisons with state-of-art methods on Glas test set and Synapse val set. We further compare our Q2A with some classic methods on Cityscapes `val` set. 'SS' and 'MS' denote single-scale and multiscale inference respectively.

segmentation benchmark datasets, i.e. Glas and Synapse (Tabs. IIIa and IIIb). We adopt the same configuration for the backbone with Sec. IV-C.1. It is obviously shown that our Q2A achieves better performance than all previous approaches. In Glas, our Q2A achieves the best performance with a ResNet-18 backbone over classic advanced medical methods (e.g. AttnUNet [24], PraNet [25]), state-of-the-art universal semantic segmentation methods with well-pre-trained backbones (e.g. SegFormer [60], SETR-PUP [61]), recent methods using strong medical Transformer backbones (e.g. Swin-UNet [35], MedT [3]), and state-of-the-art methods with hybrid backbones based on Transformer and CNNs (e.g. ConTrans [5], TransFuse [4]). In Synapse, our Q2A also achieves the best performance over many previous methods which use tailored Transformer backbones (e.g. MT-UNet [66]) or mixed backbones from Transformer and CNNs (e.g. UCTransNet [62]). Moreover, we compare our Q2A with some classic methods on semantic segmentation benchmark, Cityscapes, to illustrate its effectiveness on universal segmentation scenario (Tab. IIIc). We adopt ResNet-101 pre-trained on ImageNet as the backbone. It is shown that our Q2A outperforms some classic methods which also focus on the decoder side (e.g. CCNet [70], UPerNet [69], SFNet [14]) and methods with fancy backbone designs (e.g. SETR [61], DeepLabv3 [59]).

## V. CONCLUSION

In this paper, for the first time, we focus on the feature alignment problem that occurred in the recent emerging implicit-neural-representation-based segmentation methods in the medical area. Hence, we present Q2A, a one-step query-based feature aligning paradigm, where for each target coordinate several queries are generated and fed to an implicit fully continuous feature pyramid in order to produce the contextual features aligned to the target. To model a fully continuous feature pyramid, we present P&A, a universal partition-and-aggregate strategy for latent code acquisition of implicit neural representation, with which effective feature decoding can be achieved at arbitrary continuous resolution. Extensive experiments are conducted on two medical benchmark datasets, i.e., Glas and Synapse, and one universal dataset, i.e. Cityscapes, which all demonstrate the effectiveness of our Q2A. Importantly, our model achieves a better trade-off between segmentation accuracy and computational cost than previous methods.